\documentclass[linesnumbered,ruled,lined]{article}
\usepackage[numbers]{natbib}
\usepackage[margin = 1.5in, bottom = 1.1in, top = 1.1in]{geometry}
\usepackage{indentfirst}
\usepackage{algorithm2e}
\usepackage{amsthm}
\usepackage{amsmath}
\usepackage{graphicx}

\makeatletter

\usepackage{url}
\usepackage{amsfonts}
\usepackage{nicefrac}
\usepackage{microtype}
\usepackage{float}

\usepackage{multirow}
\usepackage{footnote}
\usepackage{tablefootnote}
\usepackage{threeparttable}

\title{Local Region Sparse Learning for Image-on-Scalar Regression}

\author{Yao Chen\footnotemark[1],\ Xiao Wang \thanks{Department of Statistics, Purdue University},
Linglong Kong \thanks{Department of Mathematical and Statistical Sciences, University of Alberta}, Hongtu Zhu \thanks{Department of Biostatistics, MD Anderson Cancer Center \& Department of Biostatistics, University of North Carolina at Chapel Hill}} 

\begin{document}

\maketitle

\begin{abstract}
Identification of regions of interest (ROI) associated with certain disease has a great impact on public health. 
%
%
Imposing sparsity of pixel values and extracting active regions simultaneously greatly complicate the image analysis. We address these challenges by introducing a novel region-selection penalty in the framework of image-on-scalar regression. Our penalty combines the Smoothly Clipped Absolute Deviation (SCAD) regularization, enforcing sparsity, and the SCAD of total variation (TV) regularization, enforcing spatial contiguity, into one group, which segments contiguous spatial regions against zero-valued background. Efficient algorithm is based on the alternative direction method of multipliers (ADMM) which decomposes the non-convex problem into two iterative optimization problems with explicit solutions. Another virtue of the proposed method is that a divide and conquer learning algorithm is developed, thereby allowing scaling to large images.
Several examples are presented and the experimental results are compared with other state-of-the-art approaches. 
\end{abstract}

\section{Introduction}

There has been significant research activity aimed at the association between image data and other scalar variables (e.g. cognitive score, diagnostic status) in the study of neurodegenerative and neuropsychiatric diseases, such as Alzheimer's disease (AD)\citep{mu11}. The growing public threat of AD
has raised the urgency to discover ROI of magnetic resonance images (MRI) that may identify subjects
at greatest risk for future cognitive decline and accelerate the testing of preventive strategies.  Machine learning methods have been developed and the penalized optimization is popular in the framework of the empirical risk minimization plus a penalty. However, spatially heterogeneous smoothness and local region selection greatly complicates the image analysis. To address these challenges, several regularization methods have been proposed to impose sparsity on both pixel values and their spatial derivatives. For instance, GraphNet \citep{eick15} combines the Lasso penalty and an $\ell_2$ penalty of image gradients, and TV-$\ell_1$ \citep{dohmatob14, meh12} uses a weighted combination of the Lasso penalty and the TV penalty. 

It is well-known that both Lasso and TV models have the inherent bias and often lead to less stable predictions \citep{tibshirani}. For example, the spatially adaptive TV model \citep{strong97} was proposed to remove the inherent bias in the TV model by utilizing a spatially varying weight
function that is inversely proportional to the magnitude of image
derivatives. It is a two-step procedure where the weight
function obtained from the first step using standard TV is then
used to guide smoothing in the second step.
It is of interest to note that, in the statistical literature, the Smoothly Clipped Absolute Deviation (SCAD) penalty \citep{fan11} has been proposed in the context of high-dimensional linear regression to address the shortcomings of Lasso (which is not consistent in variable selection). SCAD has some desired properties of the estimator such as continuity, asymptotic unbiasedness, sparsity, and the so-called oracle property (which behaves the same as when the zero coefficients are known in advance). There are a few papers on the use of SCAD for image analysis \citep{chopra10, meh12}. None of them consider the local region learning.  We will adapt the SCAD penalty for our local region selection problem in the framework of image-on-scalar regression.

In this paper, we propose a novel regularization method called SCAD2TV, which combines the SCAD regularization, enforcing sparsity, and the SCAD of TV regularization, enforcing spatial contiguity, into one group, which segments contiguous spatial regions against zero-valued background. This paper makes three main contributions:
\begin{itemize}
\item The new penalty, SCAD2TV, forces zeros on coordinates and spatial derivative jointly, which makes it easy to identify ROI for the image-on-scalar regression model. It solves the bias issue inherent in LASSO or TV methods.
\item Our proposed algorithms are based on ADMM, which decomposes a non-convex problem with the non-convex penalty into two iterative optimization problems with explicit solutions. The divide and conquer learning algorithm is also developed, thereby allowing scaling to large images.
\item Compared with GraphNet and TV-$\ell_1$, SCAD2TV has better or competitive performance in either prediction or selection errors. 
\end{itemize}



\section{Image-on-Scalar Regression and SCAD2TV}

\subsection{Image-on-Scalar Regression}
Regression models with image responses and scalar predictors are routinely encountered in many applications \citep{chen16, gold15}.  
Consider an image-on-scalar regression model with varying coefficients: $Y(s) = X^T \beta(s) + \eta(s) + \epsilon(s)$,
where $X \in \mathbb{R}^p$ is the covariate, $Y(s)\in \mathbb R$ is the image response at pixel $s\in {\cal S}$ (a 2D or 3D domain), and $\beta(s) =(\beta_1(s), \ldots, \beta_p(s))^T\in \mathbb R^p$ is the coefficient image vector.
Here $\eta(\cdot)$ is a zero-mean spatial field which characterizes the spatial correlation, and $\epsilon(\cdot)$ is the white noise with mean zero and variance $\sigma^2$. In this paper, we focus on $Y\in \mathbb R^{N\times N}$ a 2D image. Extension to 3D images is straightforward. The objective is to identify ROI in the response image which are associated with the corresponding covariate by estimating the coefficient images $\beta_1, \ldots, \beta_p$. The available data are image and covariate pairs for $n$ subjects, $(X_i, Y_i(\cdot))$, $i=1, \ldots, n$. We obtain the estimator by minimizing
\begin{equation}\label{equ:obj}
{1\over n}\sum_{i = 1}^n \sum_{s\in {\cal S}}\Big( Y_i(s) - X_i^T \beta(s)\Big)^2 + \sum_{j=1}^p\mbox{pen}(\beta_\ell),
\end{equation}
where $\mbox{pen}(\cdot)$ is a penalty function which favors estimators according to certain criteria. Our purpose is to recover nonzero active regions of $\beta_1, \ldots, \beta_p$. The main challenges are that we need to impose sparsity of pixel values and extract active regions simultaneously.



\subsection{Existing Regularizers}

\textbf{TV and SCAD}. The TV analysis plays a fundamental role in various image analyses since
the path-breaking works \citep{rudin94, rudin92}. We focus on the anisotropic version of TV.
For $\beta\in \mathbb R^{N\times N}$, define the discrete gradient $\nabla:  \mathbb R^{N\times N} \rightarrow \mathbb R^{N\times N\times 2}$ is defined by
\[
(\nabla \beta)_{jk} = \left\{\begin{array}{ll}
(\beta_{j+1,k}-\beta_{jk}, \beta_{j, k+1} - \beta_{jk}), &~~~ 1\le j, k\le N-1,\\
(0, \beta_{j, k+1} - \beta_{jk}), &~~~ j = N, 1\le k\le N-1,\\
(\beta_{j+1,k}-\beta_{jk}, 0), &~~~ 1\le j\le N-1, k=N,\\
(0, 0), &~~~ k=j=N.
\end{array}\right.
\]
The TV norm $\|\beta\|_{TV}$ is just $\|\beta\|_{TV}=\sum_{j,k}\|(\nabla \beta)_{jk}\|_1$. The isotropic induced TV norm is $\sum_{j,k}\|(\nabla \beta)_{jk}\|_2$, which is equivalent to the  anisotropic induced TV norms up to a factor of $\sqrt{2}$.

\begin{figure}[H]
  \centering
  \includegraphics[scale=0.28]{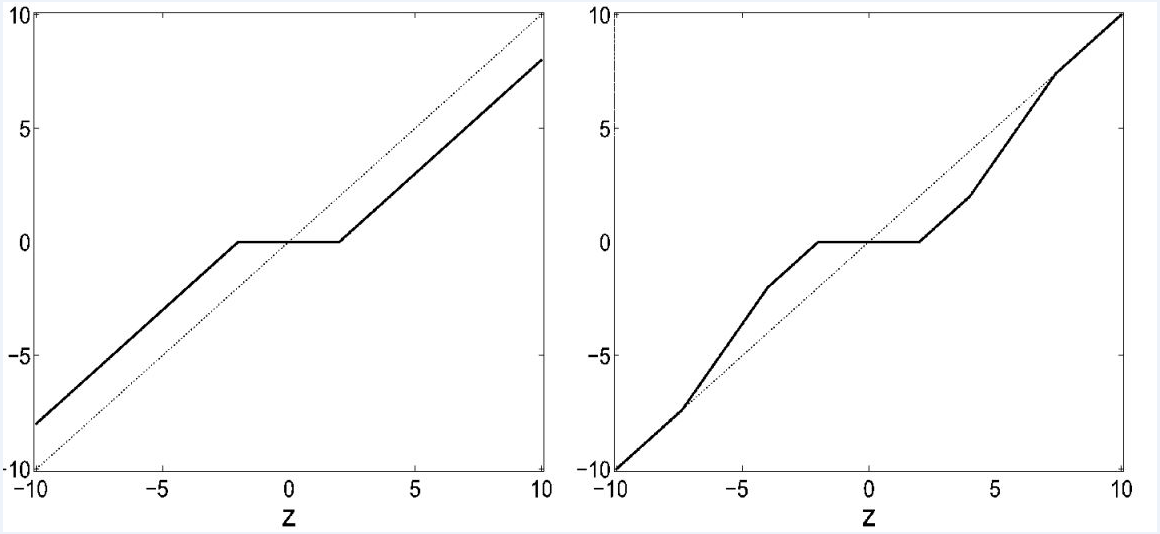}
  \caption{Thresholding function for the Lasso penalty (left) and the SCAD penalty (right) with $\varrho=1$ and $\lambda = 2$.}\label{fig:scad}
\end{figure}

The SCAD penalty $\rho_\lambda(\cdot)$ is more conveniently defined its derivative
\[
\rho_\lambda'(t) = \lambda\Big\{ I(t\le \lambda) + {(a\lambda - t)_+\over (a-1)\lambda} I(t>\lambda) \Big\}, ~~~ t>0,
\]
and $\rho_\lambda(0)=0$. We use $a=3.7$ by convention. Consider a penalized least squares problem: minimize ${\varrho\over 2}(z-\theta)^2+ \rho_\lambda(\theta)$. The solution is unique, explicit, and $\hat\theta = S_{\varrho,\lambda}(z)$, where $S_{\varrho,\lambda}$ is the thresholding function. Figure $\ref{fig:scad}$ displays the thresholding function for SCAD and the soft thresholding function for Lasso with $\varrho=1$ and $\lambda=2$. The SCAD penalty shrinks small coefficients to zero while keeping the large coefficients without shrinkage. 

\textbf{GraphNet and TV-$\ell_1$}. GraphNet and TV-$\ell_1$ have been successful applied to medical images. GraphNet is the weighted average of an $\ell_1$ penalty on all coordinates and a squared $\ell_2$ penalty on the discrete gradient, while TV-$\ell_1$ is the weighted average of an $\ell_1$ penalty and a TV penalty:
\begin{align*}
\mbox{pen}_{GN}(\beta) &= \lambda\Big( \gamma \|\nabla \beta\|_2^2 + (1-\gamma) \|\beta\|_1   \Big)  \\
\mbox{pen}_{TV-\ell_1}(\beta) & = \lambda\Big( \gamma \|\beta\|_{TV} + (1-\gamma) \|\beta\|_1   \Big).
\end{align*} 
For both penalties, $\lambda>0$ is the smoothing parameter which controls the strength of regularization and $\gamma\in [0, 1]$ is another smoothing parameter controlling the trade-off between pixel sparsity and spatial regularity.

\subsection{A New Penalty: SCAD2TV}

For each coordinate $\beta_{jk}$, the discrete gradient $(\nabla \beta)_{jk}\in \mathbb R^2$ involves three coordinate values $\beta_{jk}, \beta_{j+1, k}, \beta_{j, k+1}$. Let $\widetilde \beta_{jk} =(\beta_{jk}, \beta_{j+1, k}, \beta_{j, k+1})^T$. The SCAD2TV penalty is defined by
\begin{equation}\label{equ:pen2}
\mbox{pen}_{S2TV}(\beta) = \sum_{j, k=1}^{N}\Big\{ \gamma~ \sum_{l = 1}^2 \rho_\lambda\Big(|(\nabla \beta)_{jk,l}|\Big) + (1-\gamma)~ \sum_{l = 1}^3 \rho_\lambda\Big( |\widetilde \beta_{jk,l}| \Big)\Big\},
\end{equation}
where $\lambda>0$ and $\gamma\in [0, 1]$ are two tuning parameters, and $\rho_\lambda$ is the SCAD function. The first term in the penalty allows adaptive estimation of the coefficient image and the second one enforces sparsity on coordinate values. One may also consider the functional version of ($\ref{equ:pen2}$). After some rescaling, ($\ref{equ:pen2}$) is equivalent to 
\begin{equation}\label{equ:pen3}
\mbox{pen}_{S2TV}(\beta) = \gamma \int \rho_\lambda(|\dot{\beta}|) + (1-\gamma) \int \rho_\lambda(|\beta|).
\end{equation}
The SCAD2TV solves the bias problem inherent in the TV and Lasso models.
Note that this penalty function, unlike the $L_1$ penalty used in
Lasso, is not convex, so that ($\ref{equ:obj}$) is a non-convex objective function. We solve this problem based on the ADMM and convert it into two sub-problems with closed-form solutions. In general, ADMM has successful applications to convex problems. The behavior of ADMM applied to nonconvex problems has been a mystery. Recently,
the global convergence of ADMM in non-convex optimization is discussed in \citep{a16}, which shows that several ADMM algorithms including SCAD are guaranteed to converge.

\section{Local Region Learning by SCAD2TV}


\subsection{Algorithm based on ADMM} 

Our proposed algorithm is based on ADMM \citep{boyd11}. We may write ($\ref{equ:obj}$) as the matrix form by an abuse of notation:
\begin{equation}\label{equ:obj2}
{1\over n}\sum_{i = 1}^n \Big\| Y_i - X_i^T \beta\Big\|_2^2  + \sum_{j=1}^p\mbox{pen}_{S2TV}(\beta_j),
\end{equation}
where $Y_i\in \mathbb R^{N^2}$ is the veritorized response image for subject $i$, $X_i\in \mathbb R^{N^2\times pN^2}$ is the fixed extended design matrix related to the covariate for subject $i$, and $\beta \in \mathbb R^{pN^2}$ is the concatenated vectorized unknown coefficient image. Furthermore, one of the advantages of SCAD2TV in ($\ref{equ:pen2}$) is that we can write $\sum_{j=1}^p\mbox{pen}_{S2TV}(\beta_j)$ as $\|\rho_\lambda(D\beta)\|_1$ for a fixed $p(5(N-1)^2+6(N-1))$ by $pN^2$ matrix $D$ depending only on $\gamma$, which greatly facilitates the efficiency of our algorithm.  This fact can be easily seen since the elements involved in the $(j,  k)^{th}$ term in ($\ref{equ:pen2}$) are
\begin{align*}
((\nabla \beta)_{jk}, \widetilde{\beta}_{jk})^T = 
(\beta_{j + 1, k} - \beta_{j, k},\ \beta_{j, k + 1} - \beta_{j, k},\ \beta_{j, k},\ \beta_{j+1, k}, \ \beta_{j, k+1})^T = D_{jk} \widetilde \beta_{jk},
\end{align*}
for a fixed matrix $D_{jk}$. So $D$ is the concatenated version of $D_{jk}$.

Problem ($\ref{equ:obj2}$) is equivalent to
\begin{align*}
&\min ~~  {1\over n}\sum_{i = 1}^n \Big\| Y_i - X_i^T \beta\Big\|_2^2  + \|\rho_\lambda(\alpha)\|_1\\
&s.t. ~~~~ \alpha = D\beta.
\end{align*}
We form the augmented Lagrangian as
\[
L_\varrho(\beta, \alpha, \eta)={1\over n}\sum_{i = 1}^n \Big\| Y_i - X_i^T \beta\Big\|_F^2 + \|\rho_\lambda(\alpha)\|_1 + \eta^T(\alpha - D \beta) + {\varrho \over 2} \Big\| \alpha - D \beta \Big\|_2^2.
\]
The ADMM consists of the iterations
\begin{align}
\beta^{(t+1)} & = \arg \min_{\beta} L_\varrho(\beta, \alpha^{(t)}, \eta^{(t)})  \label{equ:beta_update} \\
\alpha^{(t+1)} & = \arg \min_{\alpha} L_\varrho(\beta^{(t+1)}, \alpha, \eta^{(t)})  \label{equ:alp_update} \\
\eta^{(t+1)} & = \eta^{(t)} + \varrho( \alpha^{(t+1)}  - D \beta^{(t+1)}). \label{equ:eta_update}
\end{align}
It should be emphasized that both ($\ref{equ:beta_update}$) and ($\ref{equ:alp_update}$) have the explicit solutions. Specifically, ($\ref{equ:beta_update}$) is a ridge regression problem and ($\ref{equ:alp_update}$) is a penalized least squares problem with the identity design matrix. The closed-form solutions for ($\ref{equ:beta_update}$) and $\ref{equ:alp_update}$ are, respectively,
\begin{align*}
\beta^{(t+1)} &= {1 \over 2} \Big({1 \over n} \mathbf{X}_i^T \mathbf{X}_i + {\varrho \over 2}D^T D \Big)^{-1} \Big({2 \over n} \sum_{i = 1}^n \mathbf{X}^T_i \mathbf{Y}_i + D^T\eta^{(t)} + \varrho D^T\alpha^{(t)}\Big) \\
\alpha^{(t+1)} & = S_{\varrho, \lambda}\Big( D\beta^{(t+1)} - {\eta^{(t)}\over\varrho}  \Big).
\end{align*}
%
%
The details of the algorithm is summarized in Algorithm $\ref{alg:admm}$.

\begin{algorithm}\label{alg:admm}
\SetKwInOut{Input}{Input}\SetKwInOut{Output}{Output}
\SetAlgoLined
\Input{Training samples $(X_1, Y_1), \ldots, (X_n, Y_n)$, tuning parameters $\lambda$, $\gamma$, $\varrho$, stopping criteria parameter $\epsilon^{pri}$ and $\epsilon^{dual}$.} 
Initialize $\beta^{(0)}$ as random uniform numbers; initialize primal and dual residuals $r^{(0)}$ and $s^{(0)}$; 

\While{$\|r^{(k)}\|_2 > \epsilon^{pri}$ and $\|s^{(k)}\|_2 > \epsilon^{dual}$}{ 
{Update $\alpha$ from ($\ref{equ:alp_update}$):} For given $\beta = \beta^{(k)}$ and $\eta=\eta^{(k)}$,  
$\alpha^{(k+1)} = S_{\varrho, \lambda}\Big( D\beta - {\eta\over\varrho}  \Big)$;

{Update $\beta$ from ($\ref{equ:beta_update}$):} For given $\alpha = \alpha^{(k+1)}$ and $\eta=\eta^{(k)}$,
$
\beta^{(k+1)} = {1 \over 2} \Big({1 \over n} \mathbf{X}_i^T\mathbf{X}_i + {\varrho \over 2}D^T D \Big)^{-1} \Big({2 \over n} \sum_{i = 1}^n \mathbf{X}^T_i \mathbf{Y}_i + D^T\eta + \varrho D^T\alpha\Big)
$;

{Update $\eta$ by ($\ref{equ:eta_update}$):} For given $\alpha = \alpha^{(k+1)}$ and $\beta = \beta^{(k+1)}$, $\eta^{(k+1)}  = \eta^{(k)} + \varrho( \alpha  - D \beta)$;

{Update $r^{(k)}$ and $s^{(k)}$:}
$
r^{(k+1)} = \alpha^{(k+1)} - D\beta^{(k+1)}, ~~~ s^{(k+1)} = \varrho D^T (\alpha^{(k + 1)} - \alpha^{(k)})$.
}

\Output{$\beta$ and $\alpha$}
\protect\caption{Local Region Learning by SCAD2TV}
\end{algorithm}
The output is either $\beta$ or $\alpha$. Note that $\alpha$ is a sparse solution and $\beta$ may not be sparse. In practice, we extract the coefficient image estimator from the output $\alpha$ to obtain the sparse estimator.

\subsection{Divide and Conquer Learning Algorithm for Large Image Size}

To address the big data issue, a divide and conquer (D\&C) algorithm is a solution by recursively breaking down a problem into two or more sub-problems of the same or related type, until these become simple enough to be solved directly. The solutions to the sub-problems are then combined to give a solution to the original problem.
In the above discussion, we assume that, for each subject $i$, all image coordinate values $Y_i\in \mathbb R^{N^2}$ are used together to infer the coefficient images $\beta$. However, in many applications $N$ may be large and such a ``batch" procedure is undesirable. In order to solve this issue, we develop a D\&C algorithm for large image size.


\begin{figure}[H]
  \centering
  \includegraphics[scale=0.18]{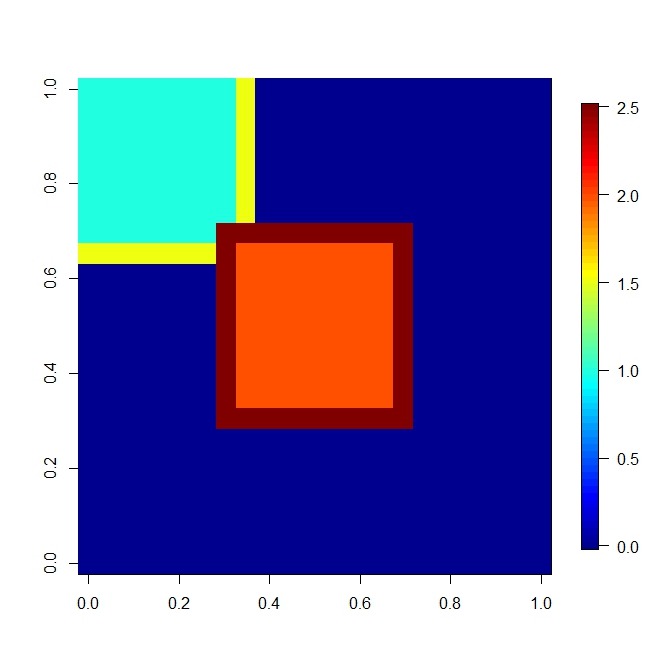}
  \caption{Partition the image into overlapped sub-images.} \label{fig:split}
\end{figure}

Image data have their intrinsic structure and we need proceed the divide step with extra caution. For example, we may just partition each image as non-overlap sub-images $Y_i = Y_{i1} \cup Y_{i2} \cup \cdots \cup Y_{iJ}$, with the data processed sequentially. Due to the TV term in SCAD2TV, we lose the boundary information for all sub-images and this will give poor estimates of the boundary for all sub-images.  
We propose to partition $Y_i$ with overlapped sub-images $Y_i = \tilde Y_{i1} \cup \tilde Y_{i2} \cup \cdots \cup \tilde Y_{iJ}$. For instance, Figure $\ref{fig:split}$ displays a $24 \times 24$ image, and it is straightforward to partition them into nine non-overlapped $8 \times 8$ sub-images. For our purpose, we extend each sub-image in four directions.
Specifically, on the left up corner, we include both the light blue part and the light yellow part, which makes our first sub-image a $9 \times 9$ image. In the center, we take the inside $8 \times 8$ orange part together with the brown part, which makes it a $10 \times 10$ image. After this partition, we obtain nine overlapped sub-images. We perform Algorithm $\ref{alg:admm}$ on each sub-image and update the coefficient images. For each update we only keep the estimate for the original  $8 \times 8$ sub-image.

The above D\&C algorithm can be executed sequentially in a single machine. This algorithm is naturally adapted for execution in multi-processor machines, especially shared-memory systems where the communication of data between processors does not need to be planned in advance, because distinct sub-problems can be executed on different processors.



\section{Empirical Results}

\subsection{Synthetic data}

We design a synthetic data example to compare the performance among three approaches: SCAD2TV, GraphNet, and TV-$\ell_1$ in terms of both prediction and selection errors.

\begin{figure}
  \centering
  \includegraphics[scale=0.34]{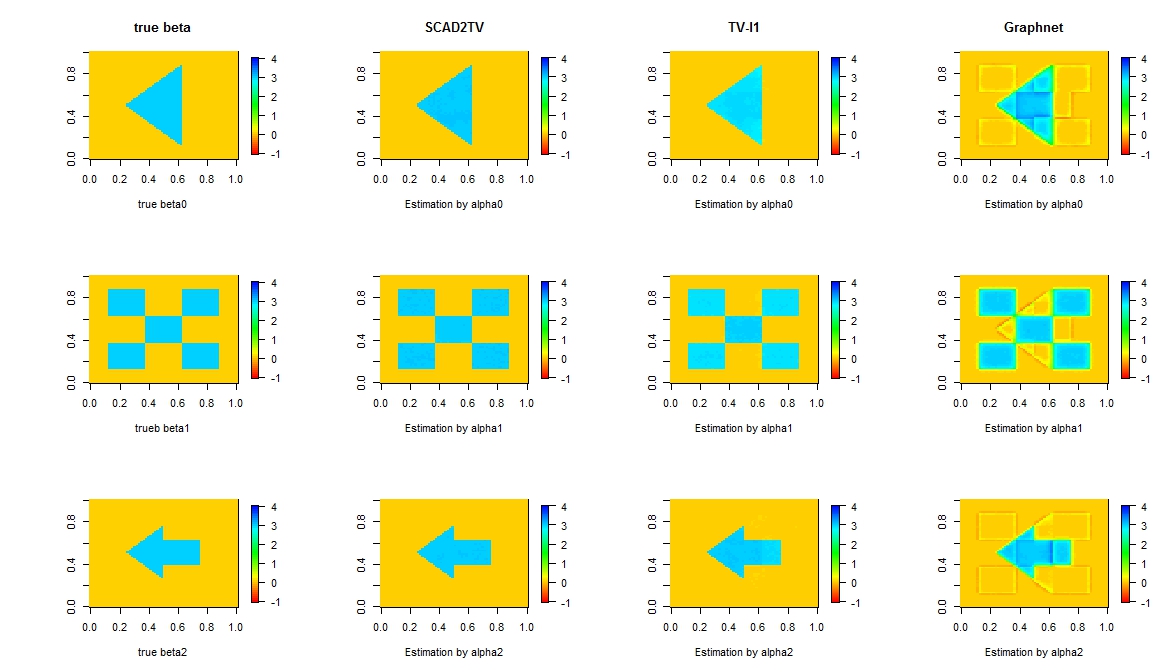}
  \caption{The coefficient images and the estimates. Left:True images; 2nd column: SCAD2TV; 3rd column: TV-$\ell_1$; Right: GraphNet.}\label{fig:bta}
\end{figure}

\textbf{Data Generation}. In our setting, $\beta=(\beta_0, \beta_1, \beta_2)$ where each $\beta_j$ is a  $64 \times 64$ image (See the left panel of Figure $\ref{fig:bta}$). The covariate is $X=(1, X_1, X_2)^T$ and each $X_j$ is generated from a uniform distribution between $0$ and $2$. The spatial field $\eta(\cdot)$ is generated from a zero mean Gaussian random field. The error process $\epsilon(\cdot)$ is the white noise with mean zero and variance $\sigma^2$. Two noise levels are adopted at $\sigma =1, 0.1$. The sample size is $n=100$.


\textbf{Applying SCAD2TV}.  In order to examine the performances of three methods, SCAD2TV, GraphNet, and TV-$\ell_1$, we have generated $100$ datasets for each setting. For each dataset, we obtain the coefficient image estimates $\hat\beta$ from these three methods. The selection rate is define as
\begin{equation}
\mbox{SR} = {1\over |S|}\sum_{s\in S} \Big( I(\beta(s) = 0, \hat{\beta}(s) = 0) + I(\beta(s) \neq 0, \hat{\beta}(s) \neq 0)\Big),
\end{equation}
and the mean squared error is defined as
\begin{equation}
\mbox{MSE} = {1 \over n  |S|} \sum_{i = 1}^n \big\|\hat{Y}_i - Y_i\big\|_2^2,
\end{equation}
where $|S|=4096$ is the total number of pixels and the $\hat Y_i$ are the predicted images. 

\textbf{Practical Consideration}. Smoothing parameters $\lambda, \gamma$ can be selected by using the K-fold cross-validation (CV). However, its computational time can be long even under current computing facilities. In our experiment,
we have tested a few different values for the tuning parameters such as $\lambda =1, 2,\ldots, 10$ and $\gamma=0.1, 0.2, \ldots, 0.9$. We find $\gamma = 0.5$ is a good balance for the estimation. The value of $\lambda$ is related to  our expectation of ROI. If the ROI has a sharp boundary and the values do not change much inside ROI, we can use a large $\lambda$. Otherwise, a smaller $\lambda$ would be preferred. We choose $\lambda = 5$, $\gamma = 0.5$ and $\rho = 1$.

\textbf{Results}. For each setting, the experiments using SCAD2TV, GraphNet, TV-$\ell_1$ are repeated $100$ times. Figure $\ref{fig:bta}$ displays the estimates of coefficient images from one realization. We note that SCAD2TV provides the solution almost exact the same as the truth. TV-$\ell_1$ can keep the sharp boundary but provide biased estimates inside the active zone. GraphNet displays blurred estimates for both active zone and zero sub-regions.  The average of the selection rates and the MSEs are reported in Table $\ref{table:sim}$. It is noted that, in terms of the selection rate, SCAD2TV performs better consistently than the other two methods. On the other hand, in terms of the prediction error, SCAD2TV and TV-$\ell_1$ are similar to each other, and GraphNet gives the highest MSE. 



\begin{table}[H]
\centering
\begin{tabular}{llllllll}
\cline{3-8}
                        &       & \multicolumn{2}{c}{SCAD2TV}        & \multicolumn{2}{c}{TV-$\ell_1$}          & \multicolumn{2}{c}{GraphNet}    \\ \cline{3-8} 
                        &       & \multicolumn{1}{c}{SR}    & \multicolumn{1}{c}{MSE}                    & \multicolumn{1}{c}{SR}    & \multicolumn{1}{c}{MSE}                    & \multicolumn{1}{c}{SR}    & \multicolumn{1}{c}{MSE}                    \\ \hline
\multicolumn{1}{l|}{\multirow{3}{*}{$\sigma$ = 1}}  	& $\beta_0$ & \multicolumn{1}{|c}{\textbf{0.820}}  & \multicolumn{1}{|c|}{\multirow{3}{*}{\textbf{1.97}}}  & \multicolumn{1}{|c|}{0.783}  & \multicolumn{1}{|l|}{\multirow{3}{*}{1.98}}  & \multicolumn{1}{|c|}{0.534}  & \multirow{3}{*}{1.98}  \\ \cline{2-3} \cline{5-5} \cline{7-7}
\multicolumn{1}{l|}{}                        		  	& $\beta_1$ & \multicolumn{1}{|c}{0.728}  & \multicolumn{1}{|c|}{}                       & \multicolumn{1}{|c|}{\textbf{0.740}}  & \multicolumn{1}{|l|}{}                       & \multicolumn{1}{|c|}{0.583}  &                        \\ \cline{2-3} \cline{5-5} \cline{7-7}
\multicolumn{1}{l|}{}                        		  	& $\beta_2$ & \multicolumn{1}{|c}{\textbf{0.677}}  & \multicolumn{1}{|c|}{}                       & \multicolumn{1}{|c|}{0.672}  & \multicolumn{1}{|l|}{}                       & \multicolumn{1}{|c|}{0.508}  &                        \\ \hline
\multicolumn{1}{l|}{\multirow{3}{*}{$\sigma$ = 0.1}} & $\beta_0$ & \multicolumn{1}{|c}{\textbf{0.9995}} & \multicolumn{1}{|c|}{\multirow{3}{*}{\textbf{0.020}}} & \multicolumn{1}{|c|}{0.9675} & \multicolumn{1}{|l|}{\multirow{3}{*}{0.024}} & \multicolumn{1}{|c|}{0.8188} & \multirow{3}{*}{0.044} \\ \cline{2-3} \cline{5-5} \cline{7-7}
\multicolumn{1}{l|}{}                        			& $\beta_1$ & \multicolumn{1}{|c}{\textbf{0.9983}} & \multicolumn{1}{|c|}{}                       & \multicolumn{1}{|c|}{0.9417} & \multicolumn{1}{|l|}{}                       & \multicolumn{1}{|c|}{0.8210} &                        \\ \cline{2-3} \cline{5-5} \cline{7-7}
\multicolumn{1}{l|}{}                        			& $\beta_2$ & \multicolumn{1}{|c}{\textbf{0.9990}} & \multicolumn{1}{|c|}{}                       & \multicolumn{1}{|c|}{0.9070} & \multicolumn{1}{|l|}{}                       & \multicolumn{1}{|c|}{0.8472} &                        \\ \hline
\\
\end{tabular} \label{table:sim}
\caption{Comparison Results of both SR and MSE for SCAD2TV, TV-$\ell_1$, and GraphNet. The bold stands for the best among three methods.}
\end{table}

\subsection{Hippocampus Data}

\textbf{Dataset}. To illustrate the usefulness of our proposed model, consider anatomical MRI data collected at
the baseline by the Alzheimer’s Disease Neuroimaging Initiative (ADNI) study, which is a large
scale multi-site study collecting clinical, imaging, and laboratory data at multiple time points
from healthy controls, individuals with amnestic mild cognitive impairment, and subjects with
Alzheimer’s disease. Given the MRI scans, hippocampal substructures were segmented with FSL FIRST \citep{pat11} and hippocampal surfaces were automatically reconstructed with the marching cube
method \citep{loren87}. We adopted a surface  fluid registration based hippocampal
subregional analysis package \citep{shi13}, which uses isothermal coodinates and 
uid registration to generate one-to-one hippocampal surface registration for surface statistics computation.

In the dataset, we have total $403$ observations. For each subject, it includes a $150 \times 100$ 2D representation of left hippocampus and $4$ covariates: gender (female=0 and male=1), age (55-92), disease status (control=0 and AD=1), and behavior score (1-36). The goal is to identify local regions of the response image associated with each covariate.

\begin{figure}[H]
  \centering
  \includegraphics[scale=0.28]{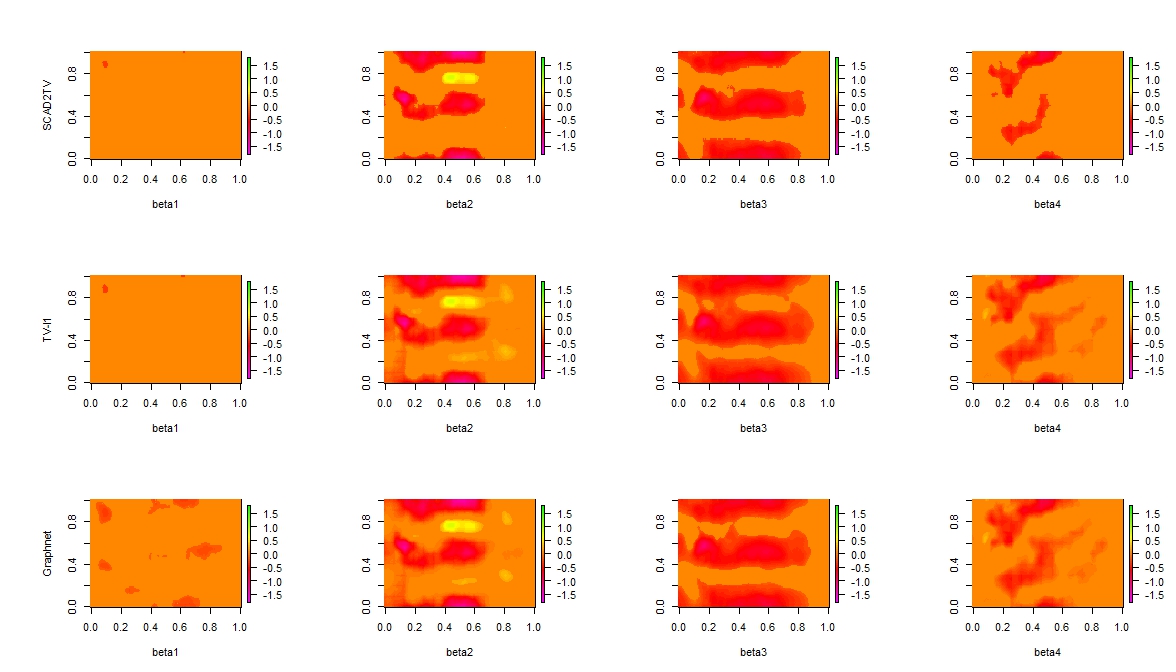}
  \caption{The estimates of coefficient images from three methods. Top: SCAD2TV; Middle: TV-$\ell_1$; Bottom: GraphNet. }\label{fig:real}
\end{figure}

\textbf{Applying SCAD2TV}. We have applied our D\&C learning algorithm to this dataset. We divide each response image into $150$ overlapped sub-images. We execute the algorithm sequentially in a single machine. The algorithm spends about 10 secs for each partition, and takes 25 minutes to get the final estimation of $\beta$'s. The estimated coefficient images are presented in the top panel of Figure $\ref{fig:real}$. 


\textbf{Results}. Our purpose is to identify the local regions where  the response hippocampus image is associated with each individual covariate. 
Among all of the covariates, we are particularly interested in the association between the response hippocampus image and the disease status (Control vs AD). From the top panel of Figure $\ref{fig:real}$, it is interesting to notice that gender has no effect on the response image, and for other three covariates SCAD2TV has been successfully identify the local active regions.

%

\begin{figure}[H]
  \centering
  \includegraphics[scale=0.15]{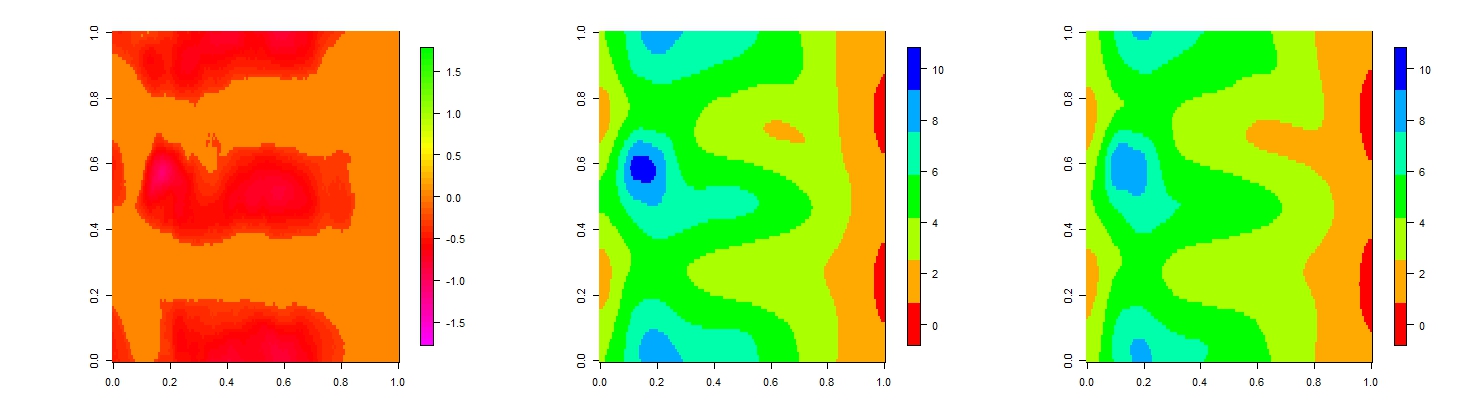}
  \caption{Left: $\hat{\beta}_3$; Middle: Mean response image for health controls; Right: Mean response image for AD.}\label{fig:real3}
\end{figure}

We take a close investigation on the coefficient image corresponding to the disease status.
The left panel of Figure $\ref{fig:real3}$ displays the estimate of $\beta_3$. The sub-regions in red indicate the active zone and the region in orange is the zero sub-region. In general, the AD patients have lower pixel values in the response hippocampus image. The right two panels are the mean response images for both health control and AD. The mean difference is consistent with our estimation of $\beta_3$.

We extract the pixels within the ROI for health controls and AD, and apply hypothesis testing to test if their difference is significant. By applying hypothesis testing on each pixel in the ROI, all of them are different between health controls and AD at the significance level $5\%$, and $99.85\%$ of the pixels in the ROI are different at the significance level $1\%$. The result justifies our ROI selection is indeed the region to differentiate between health controls and AD.

\textbf{Comparison with TV-$\ell_1$ and GraphNet}. We obtain the estimates of the coefficient images for TV-$\ell_1$ and GraphNet, which are presented in the middle and bottom panels of Figure $\ref{fig:real}$. These three methods overall detects similar regions. Both TV-$\ell_1$ and GraphNet display more blocky active regions, whereas SCAD2TV keep the active zone with sharp boundaries. We also divide our dataset into 5 parts to compare the prediction performance where each dataset contains around 80 observations. Every time we use 4 of them as the training data, and make prediction on the remaining testing data. The averages of the MSEs are computed for each methods, which are reported in Table $\ref{table:com}$. The MSEs are similar to each other and SCAD2TV displays a slightly better prediction power.

\begin{table}[H]
\centering
\begin{tabular}{cccc}
                         & SCAD2TV & TV-$\ell_1$ & Graphnet \\ \hline
\multicolumn{1}{|c|}{MSE} & \multicolumn{1}{c|}{\textbf{0.5476}} & \multicolumn{1}{c|}{0.5482} & \multicolumn{1}{|c|}{0.5491}  \\ \hline \\
\end{tabular}
\\
\caption{The MSEs for three methods}\label{table:com}
\end{table}

\section{Conclusion}

We have introduced a new region-selecting sparse non-convex penalty, SCAD2TV, which enforces large regions of zero sub-images and extracts non-zero active zones simultaneously. Efficient algorithm and the distributed algorithm have been developed.  Numerical examples are presented and the experimental results are superior or competitive with other state-of-the-art approaches such as GraphNet and TV-$\ell_1$. We have so-far focused on 2D images. It should be noted that our method works for 3D images as well. We are currently implementing our algorithm in the distributed platform such as Apache Spark. We have discussed the application for image-on-scalar regression models. This new framework may also be applied to the image clustering and image classification problems, which assume that only small regions of the images have significant effects on clustering and classification. 


\bibliography{archive}

\end{document}